# A Comprehensive Review of AI-enabled Unmanned Aerial Vehicle: Trends, Vision , and Challenges

**OSIM KUMAR PAL**\*,**MD SAKIB HOSSAIN SHOVON**†,**M. F. MRIDHA**†*(SENIOR MEMBER, IEEE)* **AND JUNGPIL SHIN**‡*(SENIOR MEMBER, IEEE)*

¹Department of Biomedical Engineering, University of Kragujevac, Kragujevac 34000, Serbia
²Department of Computer Science and Engineering, American International University-Bangladesh, Dhaka 1229, Bangladesh
³Department of Computer Science and Engineering, The University of Aizu, Aizuwakamatsu 965-8580, Japan

Corresponding author: JUNGPIL SHIN: (e-mail: jpshin@u-aizu.ac.jp)

**ABSTRACT** In recent years, the combination of artificial intelligence (AI) and unmanned aerial vehicles (UAVs) has brought about advancements in various areas. This comprehensive analysis explores the changing landscape of AI-powered UAVs and friendly computing in their applications. It covers emerging trends, futuristic visions, and the inherent challenges that come with this relationship. The study examines how AI plays a role in enabling navigation, detecting and tracking objects, monitoring wildlife, enhancing precision agriculture, facilitating rescue operations, conducting surveillance activities, and establishing communication among UAVs using environmentally conscious computing techniques. By delving intothe interaction between AI and UAVs, this analysis highlights the potential for these technologies to revolutionise industries such as agriculture, surveillance practices, disaster management strategies, and more. While envisioning possibilities, it also takes a look at ethical considerations, safety concerns, regulatory frameworks to be established, and the responsible deployment of AI-enhanced UAV systems. By consolidating insights from research endeavours in this field, this review provides an understanding of the evolving landscape of AI-powered UAVs while setting the stage for further exploration in this transformative domain.

## I. INTRODUCTION

UNMANNED aerial vehicles (UAVs) have attained popularity in purposes on surveillance, mapping, and remote sensing [1]. UAVs are becoming more popular for many reasons, such as how cheap they are to buy, how easy they are to use, how safe they are for people, and how easy it is to train people to operate them [2]. These benefits, together with their high resolution and robust tracking features, have prompted their growing use in a variety of settings.UAVs have been used for environmental monitoring, including air pollution, land surface temperature, flood danger, forest fire, road surface distress, land terrain monitoring, pedestrian traffic monitoring, and disaster evacuation [2] [3].

Drones have attained popularity in purposes on surveillance, mapping, and remote sensing [4]. UAVs are becoming more popular for many reasons, such as how cheap they are to buy, how easy they are to use, how safe they are for people, and how easy it is to train people to operate them. These benefits, together with their high resolutionand robust tracking features, have prompted their grow-ing use in a variety of settings [5]. After being utilized in geometry technologies to deliver information in a way suited for construction professionals, UAVs have become popular to handle information collecting. For example, lots of individuals are having an improved standard of life dueto sophisticated goods that can be controlled with mobile devices. Meanwhile, automobile technologies are assisting drivers by providing more up-to-date and precise information about traffic. There has been a recent uptick in the use of unmanned aerial vehicles (UAVs) in highway engineeringfor traffic monitoring and management [6].

Recent developments in computer hardware and software have made artificial intelligence (AI) a crucial component of almost every engineering-related research field [7]. Artificial intelligence (AI) is a powerful tool for tackling difficult issues for which either no clear answers exist or for which traditional methods require extensive human intervention [8]. A big difference between AI and standard cognitive algorithms is that AI can automatically extract features. This replaces expensive hand-crafted feature engineering [9].

In general, an AI job may identify anomalies, anticipate future outcomes, adapt to changing circumstances, develop understanding of complex problems requiring enormous amounts of data, and discover patterns that a person might overlook [10]. It may use and learn from the surrounding huge data to better UAV maneuvering. Comparatively to conventional optimization techniques, AI can also handle on-board resources wisely [11].Drones using AI typically have their functions fully or partially automated. With the help of AI, drone makers can use data from devices connected to the drone to collect and use data about the surroundings and how it looks [12]. AI can be used to handle drones automatically, including how they move and navigate. Several methods, including GPS monitoring, computer vision, and machine learning algorithms, can be used to accomplish this.Voice recognition, scene identification, object detection, and picture categorization are just some of the many fields where AI is making inroads, particularly when it comes to deep-learning approaches to AI [13]. Through this procedure, deep learning attempts to unearth deep characteristics in unprocessed data at various levels [14]. These characteristics are then employed to stand in for the actual world. According to projections, the UAV market would increase from USD 26.2 billion in 2022 to USD 38.3 billion by 2027, with a 7.9% CAGR over those two years. Rising acquisition of tiny drones for military applications such as ISR will boost the small drone market development throughout the projected period [15]. Despite this growing interest in UAVs, there are still a lot of restrictions. Many of the problems with UAVs, like high power/energy consumption and real-time needs, are also good things about edge computers and edge AI, like low energy consumption and low delay [16] .Deep learning and unmanned aerial vehicles (UAVs) together have the potential to revolutionize traffic monitoring in the transportation sector [17].The optimal usage of AI and deep machine learning in UAVs that are already being used and the potential future will be explored and diagnosed in this in-depth investigation, as will the large region of AI-embedded drone applications.

AI-enabled UAVs need robust computers and sensors that use a lot of energy. Energy-efficient components and algorithms are essential to green computing. Scientists can refine AI algorithms to lower processing demands and UAV power needs [18]. Renewable energy sources like solar panels or wind turbines may charge UAV batteries more sustainably. AI is essential for UAV autonomy and navigation. AI algorithms may improve UAV flight patterns, reducing fuel consumption and emissions. Researchers, industry professionals, and government organizations may collaborate to develop green computing solutions for AI-enabled UAVs. Researchers may work to develop environmentally friendly technology and techniques for better green UAVs [19] [20].

A study was undertaken of Deep Learning methods based on UAV uses. They have shown the limits of the present state of UAV development. Not sufficient attention was devoted to discussing algorithms and the potential uses of UAVs in specialized industries [14].A assessment of the ML-based UAV communication system was conducted using various algorithms. In that particular research, just few of the many potential applications of UAVs were investigated [21]. Vision-based unmanned aerial vehicle (UAV) navigation system using a variety of AI technologies were examined. Several applications of computational intelligence, including search and rescue, as well as surveillance, were addressed [22]. Rescue sharing, distribution, and trajectory design uses for smart UAV base station were surveyed. Researchers analyzed potential AI-based rescue strategies [23].Computerized vision methods based on machine learning algorithms and UAV platforms are being studied by researchers for their potential to detect and cure agricultural illnesses at an early stage.Multiple elements, including plants weeds, pests, and conditions, were taken into account in the research [24].Researchers investigated the potential for AI-based unmanned aerial vehicle (UAV) systems applications to be used for traffic management [25], as well as monitoring [26],control

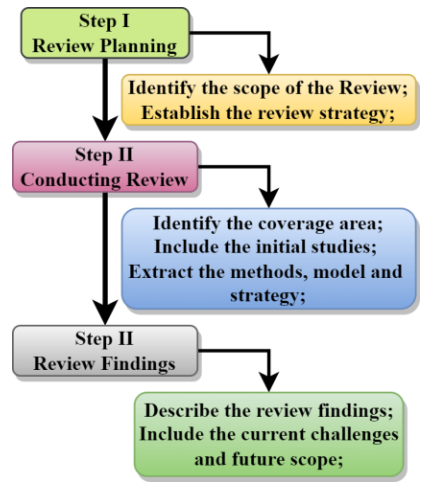

**FIGURE 1.** Outline of the Structured Review

[27], and detection of traffic [28]. Most recent survey studies concentrated their attention on one application of AI-based UAVs. The research only provided limited examples and did not cover a significant portion of the AI and ML algorithmic landscape. This review effort will evaluate the various uses of UAVs where AI is appropriate and examine all of the different learning algorithms that researchers are presently using.The investigation of AI-enabled UAVs is a captivating exploration of state-of-the-art technology. This detailed overview examines current trends, predicts a future in which UAVs will reshape several sectors, and discusses the issues that will be addressed. This discourse explores the transformative effects of AI, along with green computing and UAVs, on our society and their potential for fostering innovation and generating beneficial outcomes. This systematic

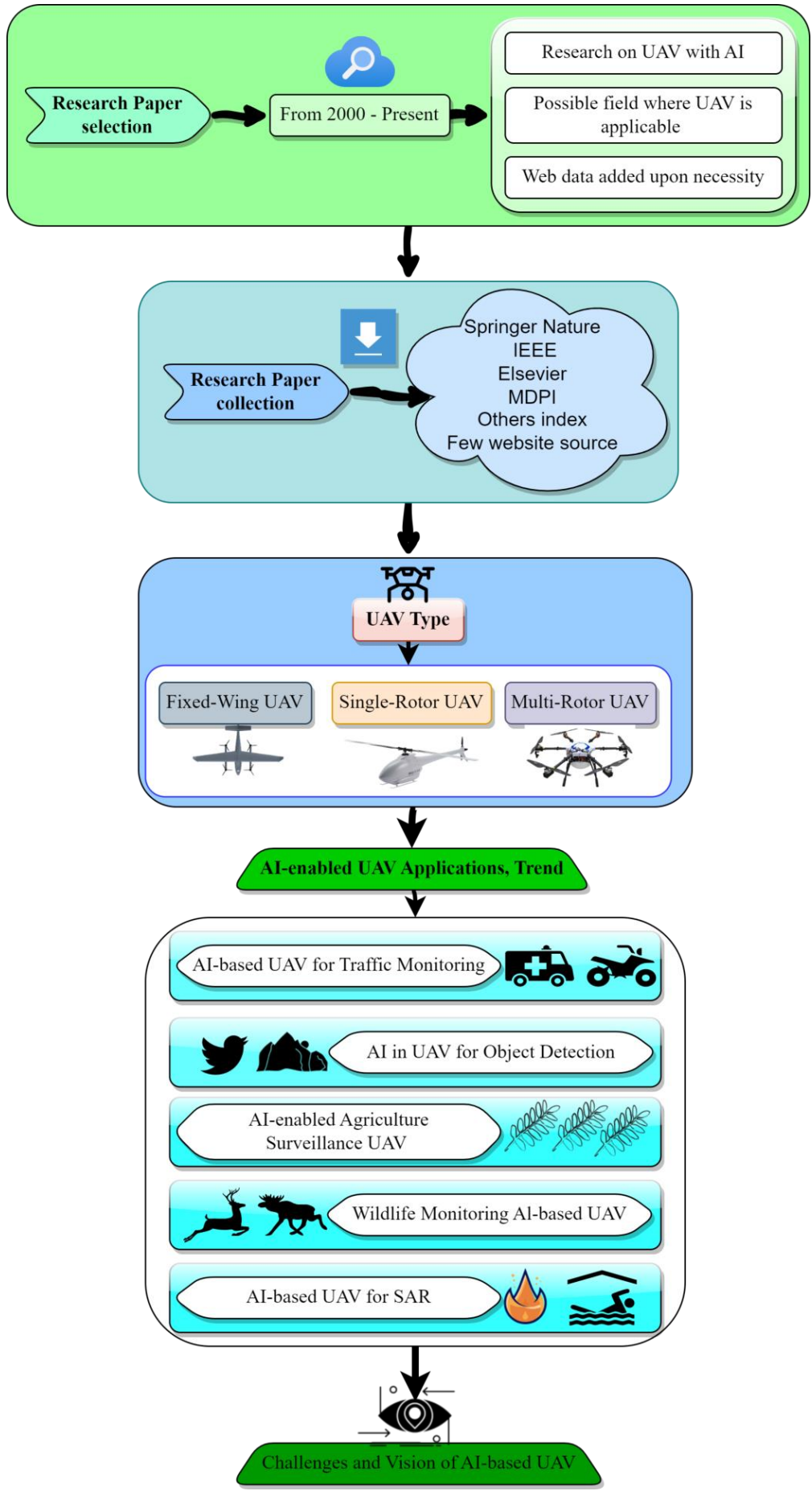

**FIGURE 2.** Review Methodology

review also includes a discussion of the possible datasets for each eligible area of UAV.The schematic representation of the review process is depicted in Figure 1.

The systematic literature review is set up in the manner described below:The research methodologies utilised for this investigation are discussed in Section 2. Types of UAV are mentioned in Section 3. Section 4 offers a thorough analysis and applications, while Section 5 lists constraints and potential future research fields. Finally, Section 6 discusses the challenges and future scope of this study.

## II. REVIEW METHODOLOGY

This section serves as an illustration of the organized and methodical approach to doing research. It includes the methods, processes, and tactics used by scholars to collect, process, and evaluate data in order to respond to certain investigations or assumptions.Figure 2 depicts the approach employed in this study. The validity, trustworthiness, and credibility of the study conclusions depend on a well established research process.

### A. PAPER SELECTION CRITERIA

As with any review, we selected and included the research work or paper based on a handful of criteria. These were the standards that we adhered to-

- The article should be directly related to UAV research or surveys.
- Artificial intelligence, deep learning, and UAV-related papers are also included in our selection.
- UAV-related research papers from several sub-fieldsare also included in our evaluation owing to extensive knowledge collection.
- We picked relatively few website data when it is statistical or genuine data that is exclusively accessible on that site.

### B. SOURCE OF INFORMATION

A successful and instructive scientific review will comprise material that has been gathered from publications that are reliable, well-respected, and well-organized. Consequently, the articles or information for this review study are collected from scientifically appreciated Scopus Journal indexes such as Springer Nature, IEEE, Elsevier, MDPI journals, Wiley, and many more. We have only acknowledged a few articles from conferences with rigorous structural criteria. A few reliable online sources compile a minimal amount of statistical data and up-to-date information.

### C. AREA OF COVERAGE

This study covers the time period from 2000 to the present. At the very beginning of the year 2000, several industries began implementing new applications for advanced UAVs. After the revolution and the growth of technology, AI allows UAV to become a necessary kit for practically every industry in order to improve surveillance, safety, security, and decision making.

## III. UAV PLATFORM TYPE

Unmanned aerial vehicles come in a wide variety, so the name "drone" is all-encompassing. It may mean either an intelligent or an autonomous aircraft. Hexacopters, quadcopters, multi-copters, and aircraft with wings all fall within this category [29]. The primary types of flying drones are:

### A. FIXED-WING UAV

A fixed-wing drone features one rigid wing that is meant to appear and function like an aircraft and provides lift instead of vertical lift rotors. Therefore, this form of drone simply requires energy to go ahead and not to maintain its airborne position [30]. They are energy-efficient as a result. Fixed-wing drones can travel farther, map much bigger areas, and stay still for a long time while keeping an eye ontheir target [31]. These drones have a higher ceiling anda greater payload capacity. Drones with fixed wings may be pricey.Flying fixed-wing drones typically requires training.

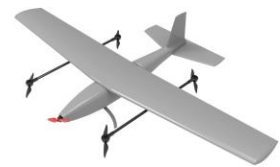

**FIGURE 3.** This is Fixed-Wing UAV model

It can be used for Aerial Mapping, Agriculture Inspection, Construction Monitoring, and numerous other applications [32].

### B. MULTI-ROTOR UAV

The simplest and least expensive method for keeping an "eye in the sky" is to use a multi-rotor drone. They also allow for more precise positioning and framing, making them ideal for aerial photography and surveillance [33].

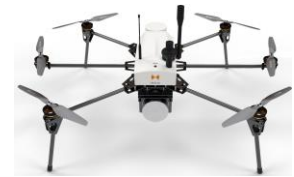

**FIGURE 4.** This is Multi-Rotor UAV model for Modern Use

Common types of multi-rotor aircraft include tri-copters (with three rotors), quadcopters (with four), hexacopters (with six), and octocopters (with eight). Quadrotors are the most prevalent multi-rotor drones. It offers superior aircraft control while in flight. Thanks to its improved maneuverability, it can go backward, forwards, sideways, and around on its axis [34]. Because of their low endurance and speed, multi-rotor drones aren't suited for extensive aerial mapping, long-term monitoring, or long-distance infrastructure inspection like highways, pipelines, and electricity lines. They are inherently inefficient and need a lot of energy to defy gravity and maintain their airborne position [1].

### C. SINGLE-ROTOR UAV

UAV with a single rotor are robust and long-lasting. They resemble helicopters in terms of construction and design.A single-rotor helicopter consists of a single rotor—similar to a large rotating wing—and a tail rotor for directional and stability control.Single-rotor helicopters are more efficient than multi-rotors, especially if they're gas-powered.Long blades spin like wings rather than propellers, making a single-rotor helicopter efficient [35]. Single-rotor drones are costly and complicated.They tremble and are less stable or tolerant of a poor landing.Because of their technical intricacy, they need frequent maintenance [36].

## IV. ARTIFICIAL INTELLIGENCE EMBEDDED UAV

UAVs and artificial intelligence (AI) are two topics that have recently attracted the interest of researchers in academia and industry [37]. Aerial drones have increased the flexibility with which operations may be carried out and activities monitored from distant areas [38]. In addition to expanding UAVs capabilities and throwing up the market to a broader variety of businesses, deploying AI and machine learninghas also helped lessen the number of obstacles that must be overcome [39]. The combination of UAVs with machine learning has led to both speedy and dependable outputs [40].Figure 5 demonstrates the present use of AI across a variety of industries, including UAV applications.

The use of unmanned aerial vehicles (UAVs) in conjunction with artificial intelligence has been shown to be advantageous for real-time monitoring, the collection and processing of data, and prediction in a variety of contexts, including cities with smarts, defence, farming, and mining [41].

### A. APPLICATIONS OF AI IN UAV FOR TRAFFIC MONITORING

The location, speed, and direction of vehicles, as well as the number of times they traverse a particular point (like a gate, a junction, or a crossing), are just a few features that UAVs can observe and determine [42] [43]. These parameters are usually determined by the UAV placed over the coverage area. Changes in the values of the parameters discussed below may be used to identify certain occurrences [44] [45].

For instance, speeding may be determined when a vehicle's speed is measured and exceeds a specified limit. On the other hand, traffic bottlenecks may be identified when the average speed of many cars drops below a specific limit [46].The camera mounted on the UAV used in a surveillance system based on UAVs gathers photos of the current traffic condition using technology associated with route planning [26] [47]. The identification system, which the UAV carries, is then automatically fed these photographs.

This identification system's primary skill is its capacity to assess traffic congestion [48]. The results of the recognition are sent to a traffic-management facility. The traffic managers may readily provide further analysis using those data [49] [50].

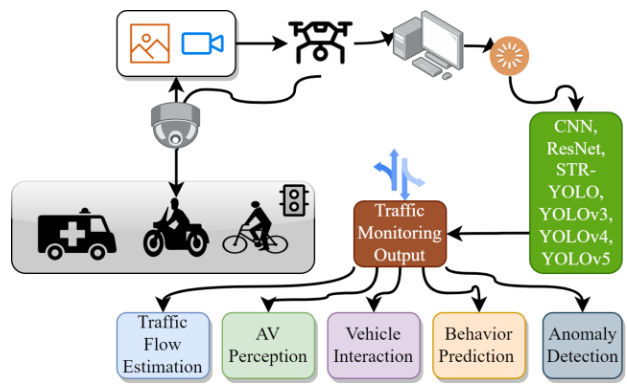

**FIGURE 5.** Traffic Monitoring UAV model for Modern Use

Figure 6 depicts the current AI applications for traffic monitoring and how UAVs function with it as an embedded system.The classification system that UAVs carry can be broken into two parts: one for extracting features and the other for recognizing characteristics [51] [52]. The recorded pictures are sent to the part of the system that pulls out the high-level details. The outcomes of the final recognition are then determined by the system component responsible for the recognition, and these findings are based on the extracted characteristics [53].

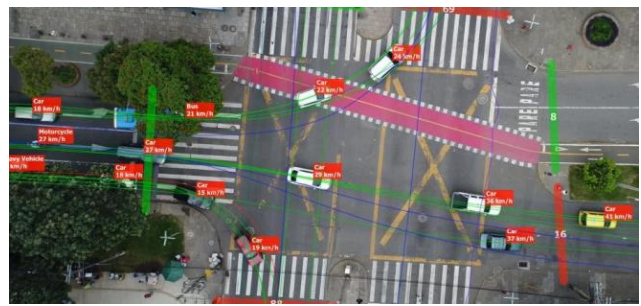

**FIGURE 6.** Traffic Monitoring using UAV

A reconstructed network that has convolutional layers is used as the fundamental architecture for the feature-extraction component of the system [54].In most cases, the additional residual block cuts down on the time required for training and the level of intricacy involved [55]. Currently, several ResNet networks (such as ResNet-50, ResNet-101, and ResNet-110) work as a learned layer for the neural network [56]. Recently, training a 110-layer ResNet with random depth has given better results than training a 110-layer ResNet with a fixed depth, and it takes a lot less time to prepare [57].Figure 7 shows how UAVs are used in real life to track and keep an eye on vehicles using AI methodsto monitor traffic.

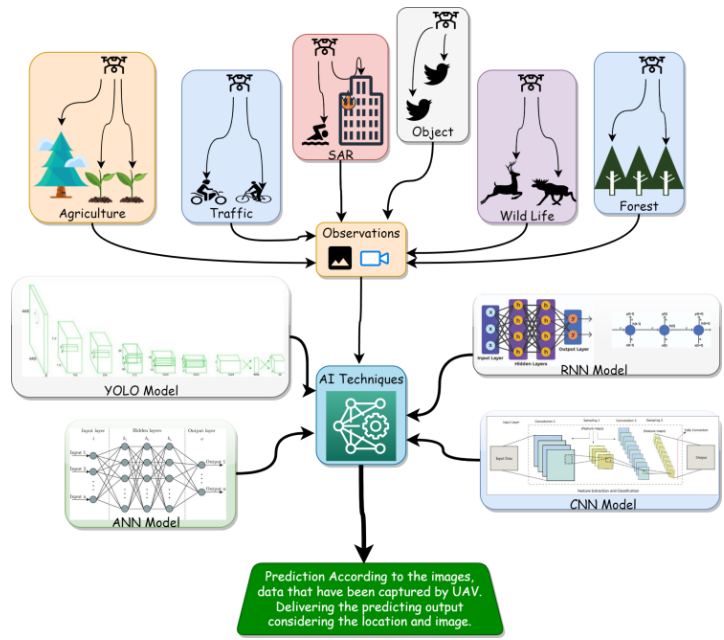

**TABLE 1.** UAV Traffic Monitoring Dataset

| Dataset | Data Type | Short Description |
|---|---|---|
| **AU-AIR [58]** | Raw Video | This dataset has 2 hours raw video of traffic monitoring along with 8 categories.GPS, time, length also mentioned in the video data. |
| **Custom & Massachusetts roads dataset [59]** | Images | These dataset contains high quality images of road,traffic sign, road conditions and vehicles |
| **Traffic Drone Data (BD) [60]** | Images | This dataset contains various road conditions data such as bus, car and other vehicles as annotated mode. |
| **Aerial Detection Dataset [61]** | High Quality Images | Dataset holds very high quality images on bus,truck, car and more. Data are split in training, test and validation. |
| **HIT-UAV [62]** | High Quality Images | Its a thermal dataset that contains over forty thousand frame of vehicles images |

YOLO is a technique that offers real-time object identification via neural networks [63]. This method is used a lot because it works quickly and correctly. It has been used to monitor traffic, people, and parking meters, among other things [64] [65].Yolo V3 switched from using Darknet-19 as the backbone network to using Darknet-53 instead [66]. Additionally, it used multi-scale estimates. Only a few academics now employ Spatial Pyramid Pooling (SPP) with Yolo V3 to identify traffic signs [67] [68].In the system that monitors traffic, the yolo algorithm is used for vehicle counting, detecting, and classifying, as well as monitoring traffic signs.

To determine the location of a vehicle, a YOLOv4-tiny model is used [69].Recently, an enhanced learning algorithm called TSR-YOLO has been designed to recognize the traffic sign in accordance to monitor the traffic [70].A system capable of performing global feature extraction with a multi-branch lightweight detection head has been developed to improve the accuracy of identifying more minor traffic signs. This method is well suited to challenging weather forecasts and environments [71].YOLOv5 (STC-YOLO) is an upgraded version of YOLO that performs better in environments with fog, snow, noise, occlusion, and blur. It is designed to monitor tiny traffic signs and vehicles [72].

The area of artificial intelligence-based UAV traffic monitoring is actively implementing green computing approaches [73]. Strategies that help reach this goal include the use of energy-efficient hardware, the integration of renewable energy sources into ground stations, the optimization of artificial intelligence (AI) algorithms to reduce the amount of computation needed, the simplification of data processing on the device itself, and the use of dynamic resource allocation. These methods aim to reduce energy consumption, decrease carbon emissions, and support environmentally sustainable aerial traffic monitoring, connecting technological advancements with environmental preservation [74].

Researchers are now employing many datasets for traffic monitoring models, and table 1 displays the most trending datasets with plausible descriptions.All systems have certain flaws and scope for development. Despite its potential bene-fits, traffic monitoring using machine learning has limitations presently being studied [75]. These algorithms may provide a few wrong results when counting vehicles since they are only partially accurate [76]. The intricacy of the environment increases the likelihood that the monitoring may give faulty results under extreme weather conditions [77]. To improve the precision and accuracy of traffic monitoring, researchers are focusing on developing these [78].

### B. AI IN UAV FOR OBJECT DETECTION

The UAV captures the camera with its lens, and machine learning and computer vision then extract the features [79]. These algorithms are capable of detecting an item's size,form, color, and ability to recognize patterns that could locate the object specifically [80]. Researchers employ sensors such as synthetic aperture radar (SAR) and light detection and ranging (LIDAR) to collect visuals, and then artifi-cial intelligence is used to extract information from those visuals to locate the item of interest. The SAR method allows researchers to improve the visual capabilities of unmanned aerial vehicles (UAVs) [81].Figure 8 illustratesthe existing AI applications used in the domain of object detection, specifically highlighting the integration of UAVs as an intelligent system. Since these techniques are simple to employ in cloudy, dark, and wet settings, they are not reliant on the climate [82].Among the portable tools for object identification is the RGB-D camera mounted to a UAV.The Parallel tracking and mapping (PTAM) method is a new development that is used with unmanned aerial vehicles (UAVs) for localization and navigation [83]. This strategy is appropriate for environments and locations that are unknown [84].One of the precise vision techniques used by UAVs to aid in object recognition is optical flow technology. It can capture images over a wide region using its long-range shooting capabilities [85].Researchers are using numerous datasets for object detection models, and table 2 shows the most popular with credible descriptions.

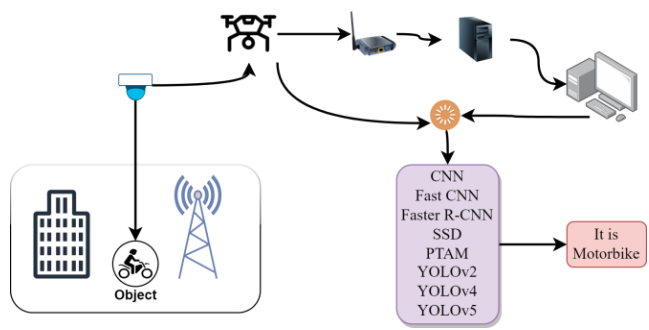

**FIGURE 8.** Object Detection UAV model for Modern Use

**TABLE 2.** UAV Object Detection Dataset

| Dataset | Data Type | Short Description |
|---|---|---|
| **UAVAT [86]** | Raw Video | This dataset has 10 hours raw video detection,multiple object tracking and single object tracking. The video is in annotated mode. |
| **UAVOD [87]** | Images | These dataset has 10 object classes such as building,ship, tower, pond, river and more. |
| **Manipal [88]** | Images | This dataset contains various small object and person data for detection |
| **Urban Zone Object detection dataset [89]** | High Quality Images | This dataset is a combination of 3 different dataset.All of the images in this dataset are well annotated. |
| **Aerial Maritime Drone Object [90]** | High Quality Images | Its has high quality images that contain boat, ship andmany more classes. |

Researchers relied on template-matching techniques in the early days of UAV objective identification . The system can identify recorded objects by comparing them to a template collection containing several thousand examples [91]. For any saved or set view, the template method works very well. However, this strategy needs to catch up in terms of large amounts of data with thousands of data categories [91] [92].At the beginning of the 2010s [93], Convolutional neural networks (CNNs) have taken over as the go-to technique for extracting the characteristics of pictures in computer vision applications [94], including picture categorization, object recognition, and semantic segmentation of photos [95] [96].

Green computing is used in AI-based UAV object iden-tification by selecting energy-efficient hardware, allocating resources intelligently, and processing data locally on the device to maximize the system's potential for reducing its overall energy consumption. This strategy lessens object de-tection missions' influence on the surrounding environment while preserving the capability to conduct effective surveil-lance. This helps to promote sustainability in aerial moni-toring activities [97].The YOLO technique was then added to UAVs for object identification and visualising. YOLO is a one-stage technique that can quickly identify a UAV-captured picture as green computing concept. Researchers have built a YOLOv2-based ROS that can communicate with

UAVs for tasks including object identification and system navigation. The method uses a simple color picture for detection [98].To advance the object detection process to the next level, researchers have identified a few potential areas for future development. An item may be located from various vantage points and angles using a UAV. This may affect the identification [99]. Another significant obstacle that must be overcome in the detecting process is the deformation of the moving item. Some items have intra-class variance, requiring additional training in ML [100]. The detection process might be positively effective if this variation is considered during training [101]. Scientists are striving to improve the UAV identification process using AI by addressing the problems mentioned earlier.

### C. AI-ENABLED AGRICULTURE SURVEILLANCE UAV SYSTEM

Agriculture has far-reaching consequences for society since it is essential in maintaining human life by providing food, shelter, and employment opportunities and supplying crucial raw materials for producing a wide variety of goods [102].UAVs have a wide variety of uses in the field of intelligent agriculture. The use of UAVs in smart agriculture allows for the evaluation of agricultural field spots, the monitoring of sunlight following the growth of crops, the diagnosis of diseases, and the administration of preventive medication for plants [103] [104].Farmers presently rely on AI-based items and applications already available for improved decision-making and earlier detection of crop conditions [105] [106]. AI enables farmers to make better decisions and more accurately assess the state of their crops [106]. The first step in intelligent farming using AI-based UAVs is harvest management, also known as yield management [107]. The authorities may also benefit from an accurate estimate of the yield since it can be used to create various strategies, including transportation needs, procurement techniques, storage facilities, and more [107] [108].The picture data is collected from the field by UAV, then artificial intelligence technology is used to analyze it [109]. This approach helps with more accurately anticipating the yield, as well as with more intelligent watering. These AI systems provide essential data on the predicted output at an early stage [109] [110].In the process of field analysis, ANN [111], CNN [112], and RNN [113] are often used to research the field picture to arrive at decisions and achieve precise positioning.

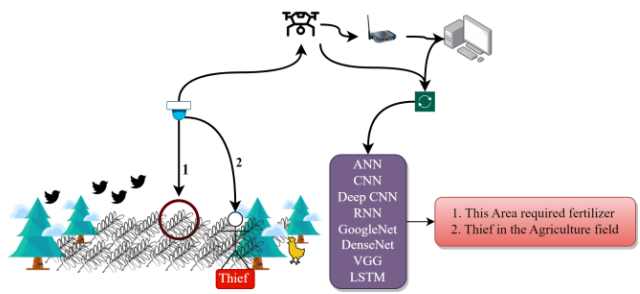

**FIGURE 9.** Agriculture Surveillance UAV model for Modern Use

Figure 9 shows how AI is already being used in farming tracking and monitoring, with a focus on how UAVs are being used as part of an intelligent system.For crop surveillance models, researchers use a lot of different datasets. Table 3 shows the most common datasets and gives reliable explanations for each.One use of UAV technology in modern farming is the diagnosis of crop diseases. Deep learning algorithms, such as CNN, Deep CNN [114], GoogleNet [94], VGG [115], DenseNet [116], and many others, are currently utilized in AI-based UAV intelligent agriculture systems for detecting disease and applying organic pesticides on the spot [117].

**TABLE 3.** UAV Agriculture Surveillance Dataset

| Dataset | Data Type | Short Description |
|---|---|---|
| **Avo-DB [118]** | Image | This dataset has RGB images and annotated images as well of a Avocado field. |
| **CoFly [119]** | Images | This dataset consists of high quality weed fields images with three different class. |
| **Paddy Field Dataset [120]** | Images | This dataset contains different types of paddy condition data for low height UAV image analysis. |
| **PlantDet [121]** | Images | This dataset hold two different types of crop image with several leaf conditions for close UAV image inspection. |
| **UAV Crop Dataset [122]** | High Quality Images | Three types of crop images from Kazakhstan with high resolution. |

When mapping land, UAVs are often utilized in place of survey drones. Survey drones powered by AI can produce high-resolution orthomosaics and comprehensive 3D models of regions that only have access to data of poor quality, that are outdated, or that do not have any data. They make it possible to construct high-accuracy cadastral maps rapidly and straightforwardly, even in locations that are difficult to access due to their complexity [123].Combining GIS with AI-enhanced drone mapping opens up whole new avenues for robots to observe and comprehend the environment [124]. Superior capacities in geographical data collection, processing, and forecasting. In the process of GIS mapping, CNN, ANN, LSTM, and Naive Bias are used extensively for image segmentation [125]. This transforms 2D photos into 3D models with high resolution for usage in GIS applications [126].

Green computing is establishing a name for itself in AI-based UAV agricultural surveillance by reducing environmental effects and maximizing efficiency in resource use. Utilizing UAV hardware that is more energy-efficient, onboard data processing, and using renewable energy sources for ground stations makes this possible. The algorithms used in machine learning are developed to reduce the amount of required computing, which helps save energy during picture analysis and data transfer [127]. The optimal use of resources may be ensured by dynamic resource allocation and intelligent scheduling, which also reduce idle periods and overall power usage. Agricultural surveillance UAVs may improve crop management while saving energy and helping sustainable farming if these green computing principles are integrated into the system [128].

### D. WILDLIFE MONITORING WITH UAV

Surveying vulnerable and invasive species to get reliable population estimates is a problematic undertaking to es-tablish the ecological balance and sustainable growth of wildlife species [129]. Intelligent UAV systems are now used to surveil forests and keep track of the animals that live there [130].UAVs that collect geo-referenced sensor data have seen rapid adoption in the last several years, particularly for ecological surveillance and animal surveying [131] [132].Figure 10 represents how AI is currently used in wildlife monitoring and tracking, with a focus on how UAVs are integrated into automated systems.Integrating green computing algorithms is pivotal in developing and implementing AI-based UAV wildlife monitoring systems. The algorithms have been specifically designed to optimize energy efficiency and mitigate environmental consequences. On-board data processing and analysis capabilities are used, reducing the need for resource-intensive data transfer to ground stations [133]. By optimizing calculations and using low-power hardware components, these algorithms contribute to the conservation of energy resources in unmanned aerial vehicles (UAVs). This, in turn, leads to an extension of flight duration and a reduction in the carbon footprint associated with monitoring activities. Successful research and protection of wildlife habitats and the promotion of sustainability in aerial surveillance activities may be achieved by integrating green computing concepts into AI-based UAV wildlife monitoring [1].

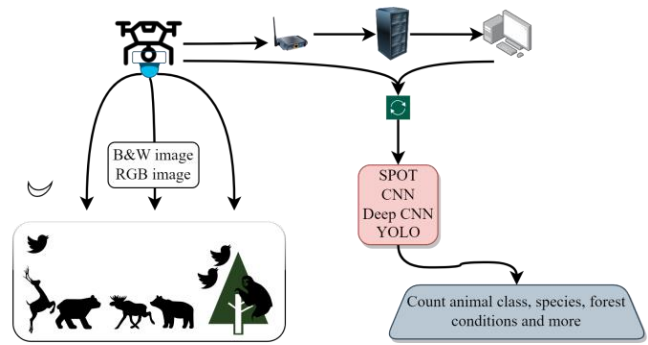

**FIGURE 10.** Wild life Monitoring

Monitoring sea turtles [134], black bears [135], big land mammals (such as elephants) [136], marine mammals (such as dugongs [137]), and birds (such as flocks of snow geese [138]), as well as providing assistance for anti-poaching activities for rhinos, are all examples of ways in which unmanned aerial vehicles (UAVs) may be used for governing wildlife [139] [140].The use of several datasets is being employed by researchers for the purpose of wildlife monitoring, counting, and surveillance models. Table 4 provides a comprehensive overview of the most often utilized datasets, accompanied by reliable and accurate descriptions.

**TABLE 4.** UAV Wildlife Monitoring Dataset

| Dataset | Data Type | Short Description |
|---|---|---|
| **Wildlife [141]** | Image | This dataset has various angle, animal and conditions of wild images. |
| **Fast Animal Detection Dataset [142]** | Images | This dataset consist of different wildlife stock,animal tracking of a wildlife reserve park. |
| **WildData [143]** | Images | This dataset has high quality wide UAV annotated images for detection. |
| **Drone Count Data [144]** | Images | This dataset hold different type of animal images for counting, analysis. |
| **UAV Aided Data [145]** | High Quality Images | Various kind of animal, trees and other species in wildlife surveillance. |

UAVs are presently used extensively for wildlife surveillance, using machine learning methods. First and foremost, UAV cameras gathered photos and video data from the forest. The picture would be in black and white, color, or RGB style for better detection and identification [146]. Regarding the data obtained from the video, the footage might be collected in its raw, night vision mode form [147].SPOT is a program that can identify poachers in longer wavelength infrared heat UAV footage instantly [148]. SPOT searches for poachers in the simulated infrared heat picture, and identifications of poachers are represented by blue rectangles [149].The system learns from examples to pinpoint the most important elements of photos taken in natural settings like forests and animals [150].

### E. RESCUE OPERATION SURVEILLANCE WITH UAV EMBEDDED AI

Embedded systems with machine learning architecture are now being taught to seek and detect persons, threats, or dangers inside a designated broad region utilizing UAVs [151].CNNs of several types, including the T(Temporal)-CNN strategy [152] [153] [154],3D-CNN [155] [156] [157], were used to identify and divide up the rescue zone. Figure 11 depicts how AI is already being employed in human and animal search and rescue, with an emphasis on the employment of UAVs as part of a smart system.

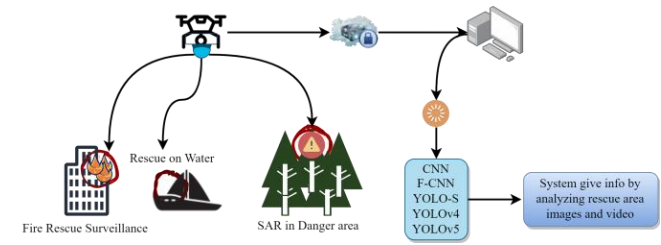

**FIGURE 11.** SAR area monitoring using AI-enabled UAV

This technique is applicable for individuals in large rivers, ponds, or perilous environments. Scientists are now using a GPU-accelerated system [158], high-altitude UAV photos [159], and feature extraction based on super-pixels [160] to locate the rescue spot. The challenge of target object recognition of various sizes is addressed by the researcher's suggested model, which uses a feature pyramid network with only one stage and a densely linked set of features [161] [162] [163].

**TABLE 5. Search and Rescue Dataset**

| Dataset | Data Type | Short Description |
|---------|-----------|------------------|
| **SARD [164]** | Image | This dataset annotated images for person and animal for rescue. |
| **Rescue Dataset [165]** | Images | This dataset has over 200 real images for search and rescue area, location. |
| **Dataset [166]** | Images | This dataset has high quality wide UAV images for rescue. |
| **SAR Human Data [167]** | Images | This dataset hold 2000 images of human action images for rescue operation. |
| **SAR Data [168]** | High Quality Images | High quality thermal image data for rescue operation and surveillance. |

The use of several datasets is being employed by researchers in the development of models for humans, small animals, item search, and rescue. Table 5 provides an overview of the most frequently utilized datasets in this domain, accompanied by reliable descriptions.First, UAV captures still images or moving video over a broad region, then analyzes using machine learning and deep learning algorithms [169] [170]. After the picture has been segmented, the algorithm will detect prospective locations where rescue may occur. This might occur in various environments, including a forest [171], a building, a flood [172] [173], a fire [174], and many more.Green computing methods are crucial for optimizing energy consumption and reducing environmental effects in AI-based UAV rescue missions. These algorithms emphasize economy by processing data locally on the UAV, negating the need for transmission, which may be taxingon system resources [175]. In this way, the UAV may save power for crucial tasks like search and rescue, even if they are meant to run on low-power technology. UAVs with artificial intelligence are more trustworthy and environmentally friendly tools for rescue missions when they adhere to green computing principles that increase their durability. This strategy is in keeping with the larger objective of ensuring the appropriate and sustainable use of technology during times of crisis [176].This lightweight version of the YOLO method may be found in UAVs as green computing concept, where it can handle various data classes at a high accuracy rate [177] [178]. YOLO is cutting-edge real-time software for analyzing images and videos [179], making it useful for SAR operations. A strategy that YOLO uses is analogous to the F(Fully)-CNN algorithm, which the SAR system utilizes. Researchers use YOLO-S [180], YOLOv4

[181], and YOLOv5 [182] on UAVs to carry out rescue operations. These techniques are readily available.

## V. CHALLENGES AND FUTURE ASPECT ON AI ENABLED UAV

Utilizing AI for UAV systems has resulted in the introduction of a multitude of innovative and resourceful solutions toan interminable array of issues [183].Drones are utilized to gather sensitive information from hazardous environments including high winds, terrible weather, heavy rain, and multi-shaded objects [184]. A vision-based unmanned aerial vehicle (UAV) navigation system using a variety of AI technologies were examined. Several applications of computational intelligence, including search and rescue, as well as surveillance, were addressed. In addition, as these systems become increasingly autonomous and interconnected, they become potential targets for criminals looking to exploit vulnerabilities. To avoid interruptions and illegal access, it is crucial to develop solid security measures and protections.

Future research in AI-based UAVs will need multidisciplinary cooperation between specialists in AI, aeronautics, ethics, law, and other fields. This cooperation is required to develop thorough regulatory frameworks that direct the use of AI-enabled UAVs while protecting against possible hazards. Research should continue to concentrate on improving AI algorithms for better navigation, judgment, and adaptability, opening the door for more advanced and trustworthy autonomous systems.Several AI algorithms are now being implemented in UAVs in order to facilitate the operation of a variety of applications. Table 6 provides an overview of the various AI algorithms currently being used in UAVs.

Table 7 lists the potential difficulties and tasks that might accelerate the development of AI-based UAVs. Also, finding new ways to use UAVs with AI, like in urban planning, environmental tracking, and crisis reaction, has the potential to change businesses and make society as a whole better. AI and UAVs working together is a growing area with much potential. Ongoing study and development will help unlock this potential and solve problems that come up along the way. As AI technology changes and new models develop,cooperation between science and technology will be critical in determining where AI-based UAVs go and how they influence our world.

## VI. REVIEW SUMMERY

"A Comprehensive Review of AI-enabled Unmanned Aerial Vehicles: Trends, Vision, and Challenges" delves into the developing scenario of AI-based UAV systems.Table 6 presents an overview of the applications of AI in several areas, with a specific focus on the current applications of UAVs.This analysis takes a look at recent developments, potential future outcomes, and the current difficulties associated with this dynamic partnership. It illustrates how AI is vital in allowing autonomous UAV capabilities, from navigation to object identification, and addresses applications such as wildlife monitoring, precision agriculture, rescue operations, and more.

**TABLE 6.** Review Summary of AI on UAV

| Applications | Algorithms uses (2000-2010) | Uses of AI algorithms | Algorithms uses (2011- Present) | Uses of AI algorithms |
|---|---|---|---|---|
| **AI based UAV for Traffic Monitoring** | CNN,RNN [54] ResNet-50 [55] ResNet-101 [56] ResNet-110 [57] | · Feature extraction of images; · Real time tracking & monitoring; · Traffic counting & survalancing; | TSR-YOLO [70] YOLOv3 [66] YOLOv4 [69] YOLOv5 [72] DarkNet-19 [67] DarkNet-53 [68] | · Identification of traffic, people, traffic place precisely; · Measure the parking space; · Vehicle counting, monitoring, detection; · Traffic sign detection in tough weather; |
| **AI in UAV for Object Detection** | CNN [94] Fast CNN [94] Fast R-CNN [95] ANN [96] | · Object detection; · Objects color, shape, size detection; · Small object detection in difficult weather; | SSD [85] PTAM [83] YOLOv2 [98] YOLOv4 [99] YOLOv5 [100] | · Identify very small object; · Detect moving & shape changing object; · Detection of small, multi-color, movable object in hazy weather; |
| **Agriculture Surveillance UAV with AI** | ANN,CNN [112] Deep CNN [114] RNN [117] | · Crop monitoring; · Crop growth & condition measure; · Farming tracking; | GoogleNet [94] DenseNet [116] VGG,LSTM [115] | · Crop disease detection; · Early stage leaf disorder prediction; · Thief, animal monitoring for crop; |
| **Wildlife Monitoring AI-based UAV** | CNN [131] Deep CNN [132] ANN [146] RNN [147] | · Animal tracking; · Wild growth monitoring; · Animal counting & survalancing; | SPOT [149] YOLO [148] | · Animal life monitoring; · Moving animal counting; · Small animal species counting, measuring; · Hunter detection; |
| **Rescue Operation Surveillance AI-based UAV** | CNN [152] F-CNN [153] T-CNN [154] 3D-CNN [155] | · Detection of person, animal; · Identifying danger, threat; · Measure rescue area condition; | YOLOvS [180] YOLOv4 [181] YOLOv5 [182] | · Identify small animal, person accurately; · Observing SAR area with possible better solution; · Assist firefighter with UAV robot; · Can applicable in forest, building, fields for SAR; |

**TABLE 7.** Challenges and Vision on AI-enabled UAV

| Application | Challenges | Future Scope |
|---|---|---|
| **AI based UAV in Traffic Monitoring** | · Large amount of video feed and real time image. · Data collection difficulties at urban environments · Detecting anomalies and triggering. · Integrating full traffic system is a major concern. | · More advanced AI for bulk number of data. · Developing fleets of autonomous UAV. · Implementing Swarm Intelligence in UAV. · Integrating Edge-Computing effectively. · Integrating predictive traffic management system. |
| **AI in UAV for Object Detection** | · Small and moving object is a major concern. · Non-annotated data creates complexity. · Adaptation to diverse environment. | · Developing Edge AI and Onboard Processing. · Domain-specific Transfer Learning. · Privacy-aware Object Detection protocol. · Developing Transferability and Generalization. |
| **AI-enabled Agriculture surveillance UAV system** | · Data management and storage might be an issue. · Crop and pest identification for numerous classes. · Interoperability with existing farming systems. · Battery Life and Flight Endurance for wide area. | · Multi-spectral and Hyperspectral Imaging. · Automated AI based Agri data analysis. · Integration with IoT devices for better analysis. · Autonomous Navigation and Dynamic Planning. |
| **Wildlife Monitoring with AI-based UAV** | · Avoiding wild disturbance might be an issue. · Adapting to difficult natural environments. · Animal movement in complex weather condition. · High resolution data collection and storage. | · Swarm intelligence integration. · Developing remote sensing technologies. · Conservation decision support improvement. · AI for numerous species identification. |
| **Rescue Operation Surveillance with UAV Embedded AI** | · Real-time situation awareness is a serious concern. · Navigating challenging environments is an issue. · Payload and Endurance. | · AI-assisted medical support. · Integrating AI for disaster prediction. · Robust design and redundancy. |

In addition, the study discusses efficient computing strategies with energy, legal issues, ethical problems, and safety precautions. This in-depth assessment may be a helpful resource for grasping the complicated nature and transformational possibilities of AI-enabled UAV technology.

## VII. CONCLUSION

The exhaustive analysis of AI-enabled UAVs reveals the dynamic nature of this technology's landscape. The study examines emerging trends that highlight the incorporation of AI, propelling unmanned aerial vehicles (UAVs) to new heights of autonomy, efficiency, and applicability across multiple industries. The review shows a picture of a future where AI-powered UAVs change businesses like tracking, detecting, espionage, transportation, and emergency management. However, the path towards this vision is fraught with obstacles, such as regulatory obstacles, ethical considerations, and technical complexities. As AI and UAV technologies continue to advance in tandem, resolving these obstacles will be essential to realizing the full potential of AI-enabled UAVs for the benefit of society.

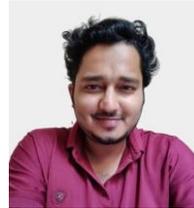

**OSIM KUMAR PAL** graduated with a Bachelor of Science in Electrical and Electronic Engineering (EEE) from American International University-Bangladesh in 2021. He is currently pursuing the Erasmus Mundus Joint Master degree in Biomedical Engineering(EMMBIOME) with the University of Kragujevac, Serbia, University of Patras, Greece and the University of Medicine and Pharmacy Grigore T. Popa, Romania, funded by the Erasmus Mundus Scholarship (2023–2025). He is working as a Research Assistant at the Advanced Machine Intelligence Research Lab (AMIRL). Previously, he joined the Department of EEE at Daffodil International University as a Teaching Apprentice Fellow. His research has been published in high-quality international venues, includ- ing academic journals and conference proceedings. His research interests include machine learning, the Internet of Things, embedded systems, robotics, informatics, and data analysis. He is now engaged in many research projects, including deep machine learning, biomedical systems, IoT and embedded systems, and so on. He can be contacted at osimkpal@gmail.com.

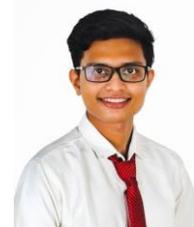

**MD SAKIB HOSSAIN SHOVON** received a B.Sc. degree in computer science and engineer-ing (CSE) with a major in information sys- tems from American International University-Bangladesh (AIUB) in 2023. He was a Research Assistant (RA) at AIUB (May 2022 – August 2022). He works at Advanced Machine Intelligence Research Lab (AMIRL) as a Research Coordinator & Lead Research Assistant (July 2022 – Continuing). In addition, he is a trainee Machine Learning (ML) Engineer under the Government Edge Project associated with HeadBlocks (28th May 2023 – ongoing). He has published many articles in prestigious journals. His research interests include Classical Machine Learning (CML), Quantum Machine Learning (QML), Federated Machine Learning (FML), Reinforcement Learning (RL), Computer Vision (CV), Natural Language Processing (NLP), Recurrent Neural Network (RNN) and Explainable AI (XAI).

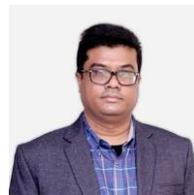

**Dr. M. F. MRIDHA**(Senior Member, IEEE) received the Ph.D. degree in AI/ML from Jahangirnagar University, in 2017. He is currently working as an Associate Professor with the Department of Computer Science, American International University-Bangladesh (AIUB). Before that, he worked as an Associate Professor and the Chairperson at the Department of CSE, Bangladesh University of Business and Technology. He also worked as the CSE Department Faculty Member at the University of Asia Pacific and as a Graduate Head, from 2012 to 2019. His research experience, within both academia and industry, results in over 120 journal and conference publications. His research work contributed to the reputed journal of Scientific Reports (Nature), Knowledge-Based Systems, Artificial Intelligence Review, IEEE ACCESS, Sensors, Cancers, and Applied Sciences. For more than ten years, he has been with the master's and undergraduate students as a supervisor of their thesis work. His research interests include artificial intelligence (AI), machine learning, deep learning, natural language processing (NLP), and big data analysis. He has served as a program committee memberin several international conferences/workshops. He served as an Associate Editor for several journals including PLOS ONE journal. He has served as a Reviewer for reputed journals and IEEE conferences like HONET, ICIEV, ICCIT, IJCCI, ICAEE, ICCAIE, ICSIPA, SCORED, ISIEA, APACE, ICOS, ISCAIE, BEIAC, ISWTA, IC3e, ISWTA, CoAST, iciVPR, ICSCT, 3ICT, and DATA21.

**JUNGPIL SHIN** (Senior Member, IEEE) received a B.Sc. in Computer Science and Statistics and an M.Sc. in Computer Science from Pusan National University, Korea, in 1990 and 1994, respectively. He received his Ph.D. in computer science and communication engineering from Kyushu University, Japan, in 1999, under a scholarship from the Japanese government (MEXT). He was an Associate Professor, a Senior Associate Professor, and a Full Professor with the School of Computer Science and Engineering, The University of Aizu, Japan, in 1999, 2004, and 2019, respectively. He has co-authored more than 300 published papers for widely cited journals and conferences. His research interests include pattern recognition, image processing, computer vision, machine learning, human-computer interaction, non-touch interfaces, human gesture recognition, automatic control, Parkinson's disease diagnosis, ADHD diagnosis, user authentication, machine intelligence, as well as handwriting analysis, recognition, and synthesis. He is a member of ACM, IEICE, IPSJ, KISS, and KIPS. He served as program chair and as a program committee member for numerous international conferences. He serves as an Editor of IEEE journals, MDPI Sensors and Electronics, and Tech Science. He serves as a reviewer for several major IEEE and SCI journals.